\documentclass[10pt,twocolumn,letterpaper]{article}

\usepackage{cvpr}              

\usepackage{graphicx}
\usepackage{nicefrac} 
\usepackage{amsmath}
\usepackage{amssymb}
\usepackage{booktabs}
\usepackage{multirow}
\usepackage{array}
\usepackage{bm}
\usepackage{utfsym}
\usepackage[accsupp]{axessibility}  

\usepackage{wasysym} 
\newcommand{\zerocircle}{\Circle} 

\usepackage[pagebackref,breaklinks,colorlinks]{hyperref}

\usepackage[capitalize]{cleveref}
\crefname{section}{Sec.}{Secs.}
\Crefname{section}{Section}{Sections}
\Crefname{table}{Table}{Tables}
\crefname{table}{Tab.}{Tabs.}
\newtheorem{Lem}{Lemma}

\definecolor{cvprblue}{rgb}{0.21,0.49,0.74}

\def\paperID{14893} 
\def\confName{CVPR}
\def\confYear{2025}
\def\tr{}

\title{A Regularization-Guided Equivariant Approach for Image Restoration}

\author{Yulu Bai$^{1,}$\footnotemark[1] \hspace{2em}  Jiahong Fu$^{1,}$\thanks{Equal contribution. \ \  \textsuperscript{$\dagger$} Corresponding author.} \hspace{2em}  Qi Xie$^{1,\dagger}$\hspace{2em}  Deyu Meng$^{1,2,3}$\\
$^{1}$  School of Mathematics and Statistics, Xi’an Jiaotong University, Xi'an, China\\
$^{2}$ Macau University of Science and Technology, Taipa, Macao \\
$^{3}$ Pengcheng Laboratory, Shenzhen, China \\
\url{https://github.com/yulu919/EQ-REG}
}

\begin{document}
\maketitle
\begin{abstract}
Equivariant and invariant deep learning models have been developed to exploit intrinsic symmetries in data, demonstrating significant effectiveness in certain scenarios.
However, these methods often suffer from limited representation accuracy and rely on strict symmetry assumptions that may not hold in practice.
These limitations pose a significant drawback for image restoration tasks, which demands high accuracy and precise symmetry representation.
To address these challenges, we propose a rotation-equivariant regularization strategy that adaptively enforces the appropriate symmetry constraints on the data while preserving the network's representational accuracy. Specifically, we introduce EQ-Reg, a regularizer designed to enhance rotation equivariance, which  innovatively extends the insights of data-augmentation-based and equivariant-based methodologies. 
This is achieved through \textbf{self-supervised} learning and the \textbf{spatial rotation and cyclic channel shift} of feature maps deduce in the equivariant framework.
Our approach firstly enables a non-strictly equivariant network suitable for image restoration, providing a simple and adaptive mechanism for adjusting equivariance based on task.
Extensive experiments across three low-level tasks demonstrate the superior accuracy and generalization capability of our method, outperforming state-of-the-art approaches. 
\end{abstract}    
\section{Introduction}
\label{sec:intro}

Various types of transformation symmetry can be observed in both local features and global semantic representations of images, including translation, rotation, reflective and scaling symmetries\cite{yosinski2015understanding}. 
Compared to fully connected networks, convolutional neural networks (CNNs) incorporate translation symmetry into their architecture. This intrinsic property has contributed to the widespread success of CNNs across various computer vision tasks, including image restoration (IR), segmentation, and recognition\cite{szegedy2015going}.
Recent advances \cite{huang2015single, he2016deep, cohen2016group, chen2022robust, kondor2018general, weiler2018learning,weiler2019general,shen2020pdo} have highlighted the crucial role of incorporating transformation symmetry priors, such as equivariant networks, in the design of network architectures.

\begin{figure}
    \centering
    \setlength{\abovecaptionskip}{0.1cm}
    \includegraphics[width=8.3cm]{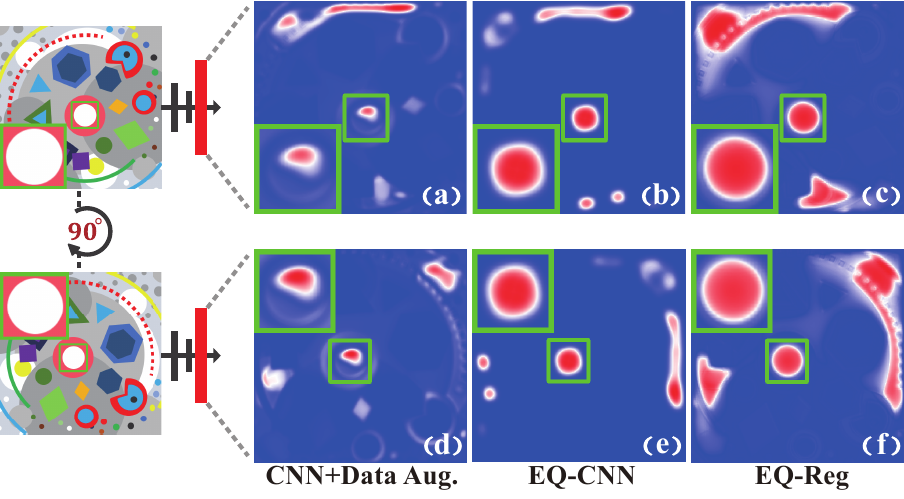}
    \vspace{-5mm}
    \caption{The feature map of the trained neural network for CNN+Data Aug., EQ-CNN\cite{xie2022fourier}, and the proposed EQ-Reg.}
    \label{fig:residual}
    \vspace{-0.6cm}
\end{figure}

Data augmentation is one of the most commonly used techniques for introducing equivariance to transformations beyond simple translation \cite{huang2015single, he2016deep, chen2022robust}. The idea is to enrich the training set with transformed samples and train networks on the augmented data set to achieve a model that is robust to transformations. A key advantage of this approach is that it enhances the network's equivariance without necessitating any structural modifications. As a result, it remains both highly feasible and effective, preserving the representational capacity while introducing transformation robustness. This is why it continues to be a widely adopted strategy in IR tasks.

However, there are also notable limitations in data-augmentation-based methods. The primary issue is that supervision imposed on the network is overly simplistic, relying solely on self-supervision at the final output. As shown in Fig.\ref{fig:method} (a), it is indeed a non-trivial task to construct regulation for feature maps in commonly used CNN, as the transformations of feature maps are unpredictable when the input image occurs transformation. While in current data-augmentation approaches, the internal feature layers of the network receive no direct regularization, resulting in limited gains in equivariance. 
In other words, networks trained with these methods only become robust to transformations, but cannot be considered truly equivariant.

Recently, incorporating transformation equivariance into network architectures has attracted significant research interest. Prominent examples in this line of work include Equivariant Convolutional Networks (EQ-CNNs) \cite{cohen2016group} and its subsequent variants \cite{weiler2018learning, weiler2019general, shen2020pdo}. 
Taking rotation equivariance as an example, the EQ-CNN framework ingeniously employs rotation and cyclic properties of convolutional kernels to achieve strictly equivariant networks. It ensures that all feature layers in the network are equivariant, meaning that as the input rotates, only \textbf{spatial rotation and channel cyclic shift} occur, with no other unpredictable changes (as shown in Fig.\ref{fig:method} (b)). More importantly, it has been theoretically proven that the EQ-CNN framework is the only method to achieve transformation equivariance\footnote{CNN is a special case that only considers translation equivariance.} \cite{kondor2018general}.

Although significant progress has been achieved in the design of equivariant networks, there are still issues that need consideration in practical applications. Particularly for IR tasks that require high representation accuracy, the most critical challenge lies in the accuracy of filter representation.

On the one hand, in the EQ-CNN framework, the convolution kernel must be transformed according to the group structure, inevitably introducing the filter parameterization techniques (as shown in Fig.\ref{fig:method} (b)). When performing filter parameterization, discrete parameters often struggle to fully capture the continuous function, leading to a degradation in representation accuracy. As a result, current EQ-CNN methods are typically limited to tasks such as classification and are less suitable for IR tasks \cite{weiler2018learning, weiler2019general, shen2020pdo}. Very recently, Xie \etal.\cite{xie2022fourier} introduced a filter Fourier-series-expansion-based parameterization approach for image super resolution, achieving an improvement in representation accuracy and verifying the importance of equivariance for IR tasks. However, the issue of representation accuracy remains unresolved, occasionally resulting in performance degradation.

On the other hand, real-world data rarely exhibit perfect transformation symmetry, implying that the strict equivariance of EQ-CNN framework may not always be appropriate. Specifically, in IR, the degradation process often violates the visibility condition \cite{beckmann2024equivariant}, leading to corrupted measurements that do not share the same symmetry priors as the underlying high-quality image. Imposing strict symmetry constraints on such inputs during neural network design can produce erroneous textures in the restored result. Recent researches have proposed multiple methods for relaxing strictly equivariant \cite{wang2022approximately,romero2022learning,van2022relaxing}. However, these approaches remain confined to the framework of group convolution. The parameter sharing techniques associated with group convolution kernels cannot circumvent the issue of diminished representation accuracy. Furthermore, there are no theoretical guarantees for these heuristic constructions. As a result, these methods cannot achieve satisfactory results in low level vision tasks.

\begin{table}[t]
    \centering
    \footnotesize %
    \captionsetup{justification=centering}
    \caption{The difference between proposed EQ-Reg and common methods. Symbol \zerocircle \hspace{1pt} represents approximate equivariance.}\vspace{-2mm}
    \begin{tabular}{>{\centering\arraybackslash}p{3.1cm}  
                    >{\centering\arraybackslash}p{1.2cm}
                    >{\centering\arraybackslash}p{1.2cm}
                    >{\centering\arraybackslash}p{1.1cm}}
    \hline
    & Data Aug. & EQ-CNN & EQ-Reg \\ 
    \hline
    Feature Map Equiv. & \usym{2717} & \usym{1F5F8} & \zerocircle \\ 
    Output Equiv. & \zerocircle & \usym{1F5F8} & \zerocircle \\ 
    Equiv. Degree Adjustable & \usym{2717} & \usym{2717} & \usym{1F5F8} \\
    Cyclic Shift of the Channel & \usym{2717} & \usym{1F5F8} & \usym{1F5F8} \\
    Representation Accuracy & Maintain & May drop & Maintain \\ 
    Beyond CNN Architecture & Easy & Hard & Easy \\ 
    \hline
    \end{tabular}
    \label{tab:compare}\vspace{-6mm}
\end{table}

To overcome the limitations of data augmentation-based and EQ-CNN-based approaches, we propose a straightforward and effective regularization-based strategy for achieving equivariance in IR tasks, currently focused on rotation-equivariant networks. The core idea is to integrate the \textbf{spatial rotation and channel cyclic shift} principles, derived from the EQ-CNN framework, into the intermediate layers of commonly used non-equivariant networks. Then, alterations in intermediate layers induced by random rotation of the input image calculated and employed as a regularization loss for self-supervision training.
The contribution of this work can be mainly summarized as follows:

\vspace{1mm}
\noindent 1) This work innovatively combines the insights of data-augmentation-based and EQ-CNN-based methods by designing rotation-equivariant regularization strategies for IR tasks. Compared to data-augmentation-based methods, our method achieves self-supervision of equivariance across all layers for the first time. Compared to EQ-CNN-based methods, the proposed method does not require modifying network architectures. Instead, we introduce a regularization term during the training process, offering a practical and effective solution. Furthermore, it opens up new avenues for designing equivariant networks beyond the traditional CNN framework. Table \ref{tab:compare} highlights the specific advantages of the proposed method over previous approaches.

\vspace{1mm}
\noindent 2) The proposed method is the first non-strictly equivariant network designed for IR tasks. Specifically, our approach introduces a simple yet effective mechanism for adaptively adjusting equivariance, making it versatile for various applications, including those involving non-strict symmetries or cases where strict equivariance is disrupted due to data degradation. As demonstrated in Figure \ref{fig:residual}, the proposed EQ-Regularization effectively adapts to capturing image symmetry priors directly from the data.

\vspace{1mm}
\noindent 3) We demonstrate the effectiveness of our approach across various IR tasks, including medical image reconstruction, image deraining, and image inpainting, as well as in image classification. Extensive experiments conducted on multiple tasks demonstrate the performance and generalizability of the proposed method, surpassing current state-of-the-art (SOTA) methods. Further analysis and visualizations validate the rationale and efficacy of our approach for broader vision tasks in real-world applications.

\vspace{-1mm}
\section{Related Work}
\label{sec:related work}

\subsection{Strict Equivariant CNNs}
Early attempts for exploiting transformation symmetry priors in images primarily rely on heuristic approaches \cite{krizhevsky2012imagenet, laptev2016ti, esteves2017polar, sohn2012learning}, with data augmentation\cite{krizhevsky2012imagenet} being the most widely used. Several attempts have been made to design end-to-end self-supervised learning frameworks that leverage equivariant priors in images \cite{chen2021equivariant, chen2022robust, zhao2024equivariant}. However, these methods  apply supervision only at the network's final output.

Recent efforts to incorporate transformation symmetry priors in images have primarily focused on embedding transformation equivariance directly into network architectures via equivariant convolution designs. Notably, GCNN \cite{cohen2016group} and HexaConv \cite{hoogeboom2018hexaconv} explicitly integrate $\pi/2$ and $\pi/3$ degree rotation equivariances into the neural network, respectively. Despite these advancements, achieving rotation equivariance for arbitrary angles remains challenging due to the reliance on discrete filter designs in these approaches. More comprehensive forms of equivariance have been explored through techniques such as interpolation \cite{zhou2017oriented, marcos2017rotation} and Gaussian resampling \cite{worrall2017harmonic}. However, in these methods, the expected equivariance cannot be theoretically guaranteed.

Current equivariant CNNs methods exploit filter parametrization technique for arbitrarily rotating filters in continuous domain. Early approaches by \cite{weiler2018learning} and \cite{weiler2019general} introduced harmonics as steerable filters to achieve exact equivariance for larger transformation groups within the continuous domain. The harmonic-based approach ensures complete rotation equivariance, making it a compelling focus for both practical applications and theoretical research. Typically, \cite{kondor2018general}, \cite{cohen2019general} provided theoretical treatment of equivariant convolution, and derive generalized convolution formulas. After that, \cite{shen2020pdo} and \cite{shen2021pdo} designed equivariance by relating convolution with partial differential operators and proposed PDO-eConv. However, these filter parameterization approaches suffer from limited expressive accuracy, which negatively impacts performance in image restoration (IR) tasks. Very recently, Xie \etal.\cite{xie2022fourier} proposed Fourier series expansion-based filter parametrization, which has relatively high expression accuracy.

\subsection{Soft Equivariant CNNs}

While symmetry constraints can be highly effective in machine learning, they may become restrictive when the data does not strictly adhere to perfect symmetry. Relaxing the rigid assumptions in equivariant networks to achieve a balance between inductive bias and expressiveness in deep learning has been the pursuit of several recent works. 

Elsayed \etal.\cite{elsayed2020revisiting} demonstrated that strict spatial invariance can be overly restrictive and that relaxing spatial weight sharing can outperform both convolutional and locally connected models. Building on this, Wang \etal.\cite{wang2022approximately} extended this weight relaxation scheme to broader symmetry groups such as rotation SO(2), scaling $\mathbb{R}_{>0}$, and Euclidean E(2). They define "approximate equivariance" explicitly and model it through a relaxed group convolutional layer.
Further, Romero \etal \cite{romero2022learning} proposed Partial-GCNN, enforcing strict equivariance on selected elements of the symmetry group. However, their method relies on constructing probability distributions of group elements and sampling from them, which both limits its applicability and adds complexity to implementation.

Finzi \etal \cite{finzi2021residual} introduced a mechanism for modeling soft equivariances by integrating equivariant and non-equivariant MLP layers. However, the high parameter count in fully connected layers makes this approach impractical for handling large-scale datasets. Similarly, Kim \etal \cite{kim2023regularizing} introduced a regularizer-based approach, using a Projection-Based Equivariance Regularizer to achieve approximate equivariances and model mixed approximate symmetry. 

In summary, most existing methods for relaxing symmetry constraints are limited to the architecture of group-equivariant convolutional networks or rely on heuristic approaches that lack theoretical guarantees. Furthermore, the effectiveness of these methods in low-level visual tasks, such as IR, remains largely unvalidated.

\begin{figure*}[ht]
    \centering
    \includegraphics[width=17cm]{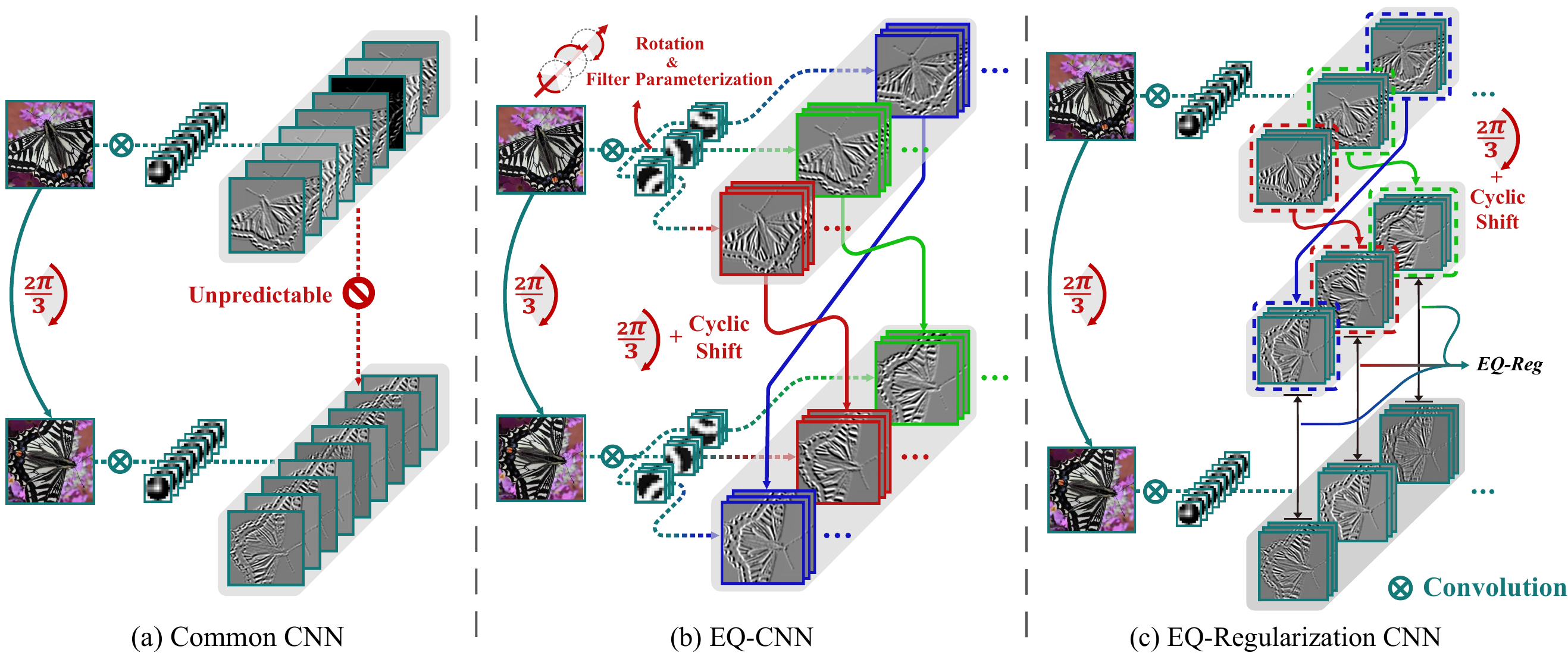}
    \vspace{-2mm}
    \caption{Illustrations of the feature maps of different CNNs when rotating the input images for $\nicefrac{2\pi}{3}$. (a) A Common CNN. (b) The rotation EQ-CNN. (c) The proposed EQ-Regularization CNN. }
    \label{fig:method}
    \vspace{-4mm}
\end{figure*}

\section{Method Framework}

In this section, we first present the foundational concepts required for constructing the equivariant regularization. Next, we describe the proposed regularizer and provide implementation details for its practical use.

\subsection{Prior Knowledge}

\noindent
{\bf Equivariance.} Equivariance of a mapping transform means that a transformation on the input will result in a predictable transformation on the output \cite{cohen2019general,shen2020pdo}. In this work, we concentrate on achieving rotation equivariance in 2D convolutions. Rotation equivariance ensures that rotating the input leads solely to a corresponding rotation in the output, without introducing any additional, unpredictable variations. Mathematically, $\Psi$ is a convolution mapping from the input feature
space to the output feature space, and $S$ is a subgroup of rotation transformations, \ie,
\begin{equation}\label{eq2}
\! \left\{\!A_k \! = \!
\begin{bmatrix}
    \cos \nicefrac{2\pi k}{t} &  \sin \nicefrac{2\pi k}{t} \\
   -\sin \nicefrac{2\pi k}{t} &  \cos \nicefrac{2\pi k}{t}
  \end{bmatrix}\big{|}  k = 0,1,\cdots,t\!-\!1
\! \right\}.
\end{equation}

\noindent
Then, $\Psi$ is equivariant with respect to $S$, if for any rotation matrix $\tilde{A}\in{S},$
\begin{equation}\label{eq3}
\Psi\left[\pi_{\tilde{A}}^I\right](I)=\pi_{\tilde{A}}^F\left[\Psi\right](I),
\end{equation}

\noindent
where \textit{I} is an input image; $\pi_{\tilde{A}}^I$ and $\pi_{\tilde{A}}^F$ denote how the transformation $\tilde{A}$ acts on input image and output features, respectively. And $[\begin{array}{c}{\cdot}\end{array}]$ denotes the composition of functions.

\vspace{1mm}
\noindent
{\bf Structure of feature maps.} For an input image \textit{I} of size $h\times w\times n_0$ ($n_0 = 1$ for grey image and $n_0 = 3$ for color image), We denote the intermediate feature map as $F$, which is a multi-channel tensor of size $h\times w\times n\times t$. Here, the third dimension corresponds to the feature channels, while the fourth dimension represents a selected rotation subgroup $S$. Following the previous work \cite{cohen2019general,shen2020pdo}, we denote the feature map corresponding to the \textit{k}-th group element as $F^{A_k}\in\mathbb{R}^{h\times w\times n}$, where $A_k$ is a rotation matrix in $S$, and also used as an index to denote a specific tensor mode in the feature map $F$.

\subsection{Proposed Regularizer}\label{sec:opt}

\noindent
\textbf{Why to Pursue High Accuracy and Adaptability?}

\noindent

Previous work has provided clear guidelines for designing neural networks to handle data with non-trivial symmetries \cite{kondor2018general}. As stated in Lemma \ref{lem1}, a feedforward network is equivariant to the group action if and only if it adheres to the convolutional operation described in Eq. (\ref{eq4}).

\vspace{-1.5mm}
\begin{Lem}\label{lem1}
A feed forward neural network $\mathcal{N}$ is equivariant to the action of a compact group G on its inputs if and only if each layer of $\mathcal{N}$ implements a generalized form of convolution derived from the following formula.
\begin{equation}\label{eq4}
    (f*g)(u)=\int_Gf(uv^{-1}) g(v) d\mu(v).
\end{equation}

\noindent
where f and $g$ are two functions $G \rightarrow \mathbb{C}$, and integration is with respect to the Haar measure $\mu$.
\end{Lem}

Therefore, the construction of a rotation-equivariant convolutional network can only proceed according to the aforementioned formulation, and the compact group $G$ as the roto-translation group $SE(2)$.

Based on Lemma \ref{lem1}, it can be proved that the current equivariant convolution is exact in the continuous domain. However, they would become approximate after necessary discretization for real-world applications. 
In this case, the latest theoretical analysis has derived the equivariant error specifically for the entire equivariant network under arbitrary rotation degrees, stated as follows \cite{fu2024rotation}:

\vspace{-1.5mm}
\begin{Lem}\label{lem2}
For an image $I$ with size $h\times w\times n_0,$ and a N-layer rotation equivariant CNN network $N_{eq}({\cdot})$, under proper conditions, the following result holds:
\begin{equation}\label{eq6}
\begin{aligned}
    |\mathrm{N}_{eq}\left[\pi_{\theta}\right](I)-\pi_{\theta}\left[\mathrm{N}_{eq}\right](I)|\leq C_{1}m^{2}+C_{2}pmt^{-1}, \\
    \forall\theta=2k\pi/t, k=1,2,\cdots,t,
\end{aligned}
\end{equation}
\noindent
where $\pi_{\theta}$ defines the rotation transformation of the input image(or on a feature map\footnote{When $\pi_{\theta}$ is the transformation on a feature map, it is combined of $\theta$ degree rotation and $k$ cyclic shift on the channels of the feature map.}), $m$ is the mesh size, $p$ is the filter size and $C_1, C_2$ is a positive constant.
\end{Lem}

The left side of Eq.(\ref{eq6}) denotes the equivariant error, i.e., the difference between results obtained when rotations are applied before and after convolution. While the equation suggests that reducing  $m$ to an infinitely small value and increasing $t$ to an infinitely large value would minimize the equivariant error to zero, achieving such extremes is impractical in real-world implementations. Consequently, the equivariant network will inevitably introduce errors, leading to reduced representation accuracy.
Furthermore, as illustrated in Fig.\ref{fig:residual} (b), with constraints such as a fixed angle selection (e.g., $t = 4$) and a finite value for $m$, the equivariant network learns a strictly symmetry constrained to four angles. This results in the circular features in Fig.\ref{fig:residual} (b) and (e) appearing nearly square.
\vspace{1mm}

\noindent
\textbf{How to Achieve High Accuracy and Adaptability?}
\vspace{1mm}

\noindent
In equivariant network, the rotation of input image $I \in \mathbb{R}^{h\times w \times n_0}$ will cause the rotation of feature map $F \in \mathbb{R}^{h\times w \times n}$ in convolution. In addition to rotating in the spatial dimension, feature $F$ will naturally accompanied by a pattern transformation respect to the rotation group in $S$ due to the property of the group equivariant network. Specifically, the rotation of the feature map would be consistent with spatial rotation on the first two dimensions and cyclically shifting along the final dimensions, as illustrated in Fig.\ref{fig:method} (b). Formally, for $\forall \tilde{A}\in{S},$ we have 
\begin{equation}\label{eq5}
\pi_{\tilde{A}}^F(F) \! = \! \left[ \pi_{\tilde{A}}^F \! \left( \! F^{\tilde{A}^{-1}\!A_1} \! \right)\!, \! \pi_{\tilde{A}}^F \! \left( \! F^{\tilde{A}^{-1}\!A_2} \! \right)\!,\!\cdots\!, \! \pi_{\tilde{A}}^F \! \left( \! F^{\tilde{A}^{-1}\!A_t} \! \right)\!\right] \! ,
\end{equation}

\noindent
where $F^{A_k}, k=1,2,..,t$, are tensors of size $h\times w\times n,$ which is viewed as a n-channel image when performing spatially rotation $\pi_{\tilde{A}}^{I}$.

To overcome the limitations imposed by representation accuracy errors of equivariant network and effectively incorporate rotational symmetry priors in images, we draw inspiration from the inherent rotational and cyclic shift properties of feature map channels in group equivariant convolution. 
We introduce a regularization term, as defined in Eq.(\ref{layer_loss}), which constrains feature maps at each convolutional layer, thereby enhancing the network equivariance.

\begin{equation}
    L_{layer}=\|\pi_{\tilde{A}}^{F}(\phi^{(l)}(I))-\phi^{(l)}\left(\pi_{\tilde{A}}^{I}(I)\right)\|_{F}^2,
    \label{layer_loss}
\end{equation}
\noindent
where, $\phi^{(l)}$ is the first $l$ layers of the convolution network. $\tilde{A}$ is a rotation matrix with randomly selected angles in $\left\{2k\pi/t|k=1,2,\cdots t\right\}$.

For the first term on the right side of Eq.(\ref{layer_loss}), denoted $\phi^{(l)}(I):= F_l$, $\pi_{\tilde{A}}^F(F_l)$ is defined in Eq.(\ref{eq5}). For the second term on the right side of Eq.(\ref{layer_loss}), denoted $\pi_{\tilde{A}}^{I}(I):= r(I)$, we have:
\begin{equation}\label{rot_then_cnn}
    \vspace{-1mm}
    \phi^{(l)}\left(r(I)\right) = \left[ r(F)^{A_1}, r(F)^{A_2}, \cdots, r(F)^{A_t} \right].
\end{equation}

According to above formula, we calculate the error between the feature maps obtained by rotating the image before convolution and those obtained by convolving the image first and then rotating it, followed by cyclic shifting on channels, as shown in Fig.\ref{fig:method} (c). Instead of using equivariant convolution architecture which would lead to representation accuracy errors, we still use common convolution layer which can maintain the high accuracy. Our goal is to guide the feature maps after each convolution layer of the image so that they can approach the strictly equivariant feature maps in group equivariant convolution networks. 

Then, the whole regularization term of the network be formulated as:
\vspace{-2mm}
\begin{equation}
    L_{equi}=\mathbb{E}_{\tilde A}\left(\sum_{l}\left\|\pi_{\tilde A}^{F}(\phi^{(l)}(I))-\phi^{(l)}(\pi_{\tilde A}^{I}(I))\right\|_{F}^{2}\right).
    \label{all_loss}
    \vspace{-2mm}
\end{equation}
We utilize this regularization to guide training, allowing the model to learn an adaptive symmetry prior from data.

CNNs trained with the regularization term, \ie Eq. (\ref{all_loss}), are not constrained by the assumption of \textbf{strict rotational symmetry} prior imposed by rotation equivariant convolution. Instead, they learn a rotational symmetry prior that adapts to the distribution of real-world degraded image. Furthermore, unlike other methods with relaxed equivariant constraints, our method achieves high-precision image restoration with a solid theoretical foundation.

\noindent
{\bf Implementation Details.} 
Since the proposed method is a basic convolution module, it can be adopted to arbitrary network in a plug and play manner. For an input image $I \in \mathbb{R}^{h \times w \times n}$, a random rotation transformation $\tilde{A}$ from the rotation group is applied to $I$, resulting in $\pi_{\tilde{A}}^I (I)$. Both the original image $I$ and $\pi_{\tilde{A}}^I (I)$ are simultaneously passed through the first layer $\phi^{(1)} (\cdot)$ to obtain the output features $\phi^{(1)} (I)$ and $\phi^{(1)} \left( \pi_{\tilde{A}}^R (I) \right)$. As shown in Fig.\ref{fig:method}(c), we perform spatial rotation and group channel cyclic shift on the feature $\phi^{(1)} (I)$ according to the transformation $\tilde{A}$ to obtain $\pi_{\tilde{A}}^{F} \left( \phi^{(1)} (I) \right)$. 
Subsequently, the loss $L_{layer}^{(1)}$ for the first layer is computed using Eq.~(\ref{layer_loss}). The process is then iteratively applied to the subsequent layers, culminating in the computation of the overall equivariant loss $L_{equi}$ for the entire network.

\begin{table*}
    \vspace{-2mm}
    \footnotesize
    \renewcommand\arraystretch{0.95}
      \centering
            \caption{Average PSNR/SSIM of different competing methods on synthesized DeepLesion~\cite{yan2018deeplesion}.} \vspace{-2mm}
               \setlength{\tabcolsep}{11pt}
            \begin{tabular}{lcccccc}
    \toprule
Method & \multicolumn{5}{c}{ Large Metal \quad \quad   \quad\quad  $\longrightarrow$ \quad \quad Medium Metal \quad \quad $\longrightarrow$    \quad   \quad        Small Metal}                & Average      \\
\cmidrule(r){1-1} \cmidrule(r){2-6} \cmidrule(r){7-7}
Input & 24.12 / 0.6761 & 26.13 / 0.7471 & 27.75 / 0.7659 & 28.53 / 0.7964 & 28.78 / 0.8076 & 27.06 / 0.7586             \\
LI~\cite{kalender1987reduction} & 27.21 / 0.8920 & 28.31 / 0.9185 & 29.86 / 0.9464 & 30.40 / 0.9555 & 30.57 / 0.9608 & 29.27 / 0.9347   \\
NMAR~\cite{meyer2010normalized} & 27.66 / 0.9114 & 28.81 / 0.9373 & 29.69 / 0.9465 & 30.44 / 0.9591 & 30.79 / 0.9669 & 29.48 / 0.9442     \\
CNNMAR~\cite{zhang2018convolutional} & 28.92 / 0.9433 & 29.89 / 0.9588 & 30.84 / 0.9706 & 31.11 / 0.9743 & 31.14 / 0.9752 & 30.38 / 0.9644   \\
DuDoNet~\cite{lin2019dudonet} & 29.87 / 0.9723 & 30.60 / 0.9786 & 31.46 / 0.9839 & 31.85 / 0.9858 & 31.91 / 0.9862 & 31.14 / 0.9814   \\
DSCMAR~\cite{yu2020deep} & 34.04 / 0.9343 & 33.10 / 0.9362 & 33.37 / 0.9384 & 32.75 / 0.9393 & 32.77 / 0.9395 & 33.21 / 0.9375 \\
InDuDoNet~\cite{wang2021indudonet}& {36.74 / 0.9742} &{39.32 / 0.9893} & {41.86 / 0.9944} &{44.47 / 0.9948} & {45.01 / 0.9958} & {41.48 / 0.9897}\\
\cmidrule(r){1-1} \cmidrule(r){2-6} \cmidrule(r){7-7}
ACDNet~\cite{wang2021dicd} & {37.84 / 0.9894} & {39.74 / 0.9928} & {41.86 / 0.9950} & {43.24 / 0.9960} & {43.96 / 0.9964} & {41.33 / 0.9939} \\
\tr{ACDNet-$partial$ } & {34.33 / 0.9591} & {37.21 / 0.9754} & {39.02 / 0.9828} & {41.00 / 0.9854} & {41.28 / 0.9866} & {38.57 / 0.9978} \\
\tr{ACDNet-$pdoe$ }&\tr{35.09 / 0.9773}&\tr{38.22 / 0.9880}&\tr{40.43 / 0.9928}&\tr{43.26 / 0.9950}&\tr{43.75 / 0.9955}&\tr{40.15} / {0.9897}\\
\tr{ACDNet-$e2cnn$ }&\tr{37.00 / 0.9861}&\tr{39.04 / 0.9908}&\tr{41.29 / 0.9940}&\tr{43.40 / 0.9955}&\tr{43.35 / 0.9959}&\tr{40.81} / {0.9925}\\
\tr{ACDNet-$gcnn$ }&\tr{37.70 / 0.9878}&\tr{40.14 / 0.9926}&\tr{41.97 / 0.9948}&\tr{43.75 / 0.9962}&\tr{43.86 / 0.9964}&\tr{41.48} / {0.9935}\\
\tr{ACDNet-$fconv$}&\tr{38.60 / 0.9888}&\tr{40.20 / 0.9928}&\tr{\textbf{42.24} / 0.9950}&\tr{\textbf{44.42} / 0.9961}&\tr{44.58 / 0.9965}&\tr{42.01 / 0.9938} \\
\tr{ACDNet-$reg$}&\tr{\textbf{38.99} / \textbf{0.9899}}&\tr{\textbf{40.60} / \textbf{0.9932}}&\tr{42.13 / \textbf{0.9954}}&\tr{43.88 / \textbf{0.9965}}&\tr{\textbf{44.76} / \textbf{0.9967}}&\tr{\textbf{42.07} / \textbf{0.9943}} \\

\cmidrule(r){1-1} \cmidrule(r){2-6} \cmidrule(r){7-7}
DICDNet \cite{wang2021dicd} & {38.88 / 0.9895}  & {39.86 / 0.9923}  & {42.85 / 0.9951} & {44.61 / 0.9957} & {45.74 / 0.9965} & {42.38 / 0.9938} \\
\tr{DICDNet-$partial$ }&\tr{37.46 / 0.9861}&\tr{39.85 / 0.9915}&\tr{41.94 / 0.9941}&\tr{44.77 / 0.9954}&\tr{44.90 / 0.9960}&\tr{41.79} / {0.9926}\\
\tr{DICDNet-$pdoe$ }&\tr{35.43 / 0.9774}&\tr{38.48 / 0.9873}&\tr{41.03 / 0.9924}&\tr{43.79 / 0.9943}&\tr{44.32 / 0.9950}&\tr{40.61} / {0.9893}\\
\tr{DICDNet-$e2cnn$ }&\tr{37.30 / 0.9854}&\tr{39.76 / 0.9910}&\tr{42.08 / 0.9939}&\tr{44.95 / 0.9954}&\tr{45.13 / 0.9957}&\tr{41.84} / {0.9923}\\
\tr{DICDNet-$gcnn$ }&\tr{38.18 / 0.9879}&\tr{39.81 / 0.9919}&\tr{42.33 / 0.9946}&\tr{45.03 / 0.9959}&\tr{45.41 / 0.9963}&\tr{42.15} / {0.9933}\\
\tr{DICDNet-$fconv$ }&\tr{38.76 / 0.9886 }&\tr{39.96 / 0.9922}&\tr{42.91 / 0.9950}&\tr{44.99 / 0.9958}&\tr{45.72 / 0.9964}&\tr{42.47 / 0.9936}\\
\tr{DICDNet-$reg$}&\tr{\textbf{39.23} / \textbf{0.9901}}&\tr{\textbf{40.23} / \textbf{0.9927}}&\tr{\textbf{43.47} / \textbf{0.9955}}&\tr{\textbf{46.08} / \textbf{0.9966}}&\tr{\textbf{46.20} / \textbf{0.9968}}&\tr{\textbf{43.04} / \textbf{0.9943}} \\
\cmidrule(r){1-1} \cmidrule(r){2-6} \cmidrule(r){7-7}
OSCNet \cite{wang2022OSC} & {39.04 / 0.9895} & {40.09 / 0.9924} & {43.12 / 0.9952} & {44.93 / 0.9957} & {45.97 / 0.9965} & 42.63 / 0.9939 \\
\tr{OSCNet-$partial$ }&\tr{37.33 / 0.9859}&\tr{40.02 / 0.9918}&\tr{42.28 / 0.9943}&\tr{44.99 / 0.9956}&\tr{45.30 / 0.9960}&\tr{41.98} / {0.9927}\\
\tr{OSCNet-$pdoe$ }&\tr{35.51 / 0.9777}&\tr{38.29 / 0.9874}&\tr{40.84 / 0.9921}&\tr{43.50 / 0.9941}&\tr{43.90 / 0.9948}&\tr{40.41} / {0.9892}\\
\tr{OSCNet-$e2cnn$ }&\tr{37.17 / 0.9847}&\tr{39.27 / 0.9907}&\tr{42.08 / 0.9940}&\tr{44.44 / 0.9951}&\tr{44.93 / 0.9957}&\tr{41.58} / {0.9920}\\
\tr{OSCNet-$gcnn$ }&\tr{38.51 / 0.9885}&\tr{40.03 / 0.9923}&\tr{42.68 / 0.9949}&\tr{45.42 / 0.9962}&\tr{45.50 / 0.9964}&\tr{42.43} / {0.9937}\\
\tr{OSCNet-$fconv$ }&\tr{\textbf{39.14} / 0.9895}&\tr{40.35 / 0.9927}&\tr{42.94 / 0.9952}&\tr{45.63 / 0.9963}&\tr{45.83 / 0.9965}&\tr{42.78} / {0.9940}\\
\tr{OSCNet-$reg$}&\tr{38.92 / \textbf{0.9898}}&\tr{\textbf{40.79} / \textbf{0.9932}}&\tr{\textbf{43.33} / \textbf{0.9954}}&\tr{\textbf{45.91} / \textbf{0.9965}}&\tr{\textbf{46.02} / \textbf{0.9967}}&\tr{\textbf{43.00} / \textbf{0.9943}} \\
    \bottomrule
\end{tabular}
    \vspace{-0.2cm}
      \label{tab:tabmar}
\end{table*}

\begin{figure*}[ht]
    \centering
    \includegraphics[width=17cm]{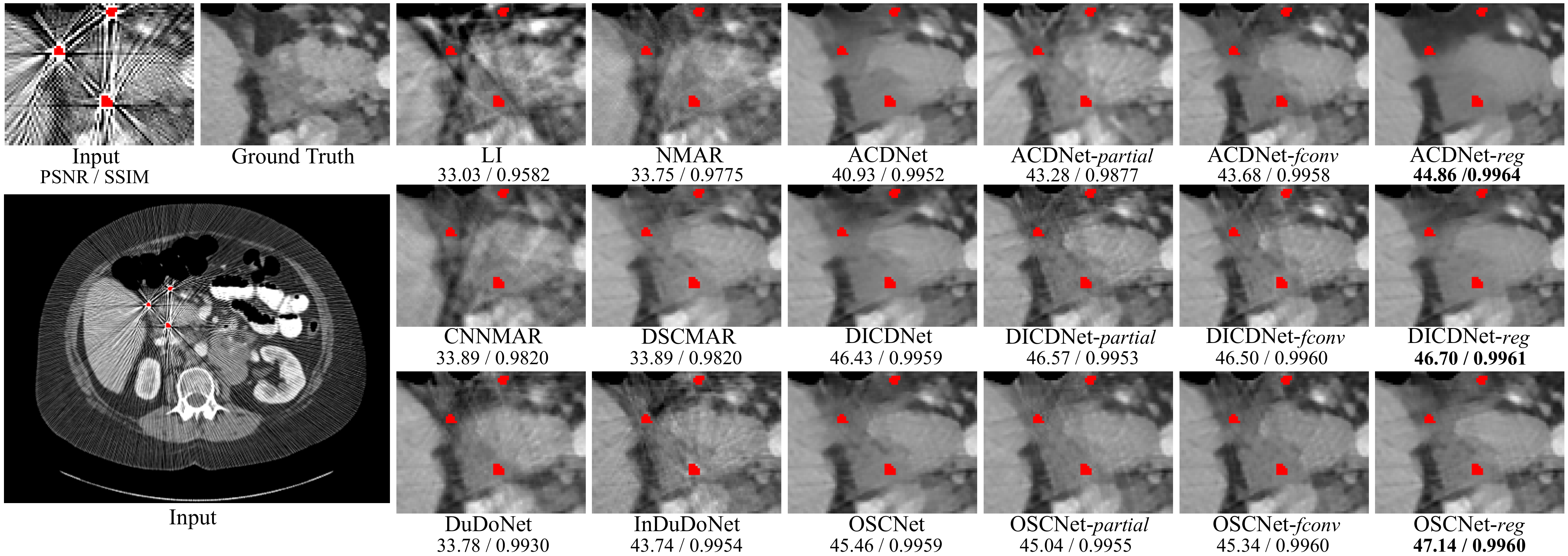}
    \vspace{-2.5mm}
    \caption{Performance comparison on a typical metal-corrupted CT image from the synthesized DeepLesion\cite{yan2018deeplesion}. The red pixels stand for metallic implants.}
    \label{fig:mar}
    \vspace{-4mm}
\end{figure*}

\begin{table*}
\vspace{-3mm}
\footnotesize
\renewcommand\arraystretch{1}
  \centering
    \caption{Average PSNR/SSIM of different competing methods on four benchmark datasets.}\vspace{-2mm}
    \setlength{\tabcolsep}{11pt}
  \begin{tabular}{ccccccccc}
    \toprule
  \multirow{2}{*}{Method} & \multicolumn{2}{c}{Rain100L \cite{yang2019joint}} & \multicolumn{2}{c@{}}{Rain100H \cite{yang2019joint}} & \multicolumn{2}{c}{Rain1400 \cite{fu2017removing}} & \multicolumn{2}{c@{}}{Rain12 \cite{li2016rain}}\\
\cmidrule(r){2-3}\cmidrule(r){4-5}\cmidrule(r){6-7}\cmidrule(r){8-9}
   & PSNR & SSIM & PSNR & SSIM  & PSNR & SSIM & PSNR & SSIM\\
    \midrule
  Input & 26.90 & 0.8384 & 13.56 & 0.3709 & 25.24 & 0.8097 & 30.14 & 0.8555\\

  DSC\cite{luo2015removing} & 27.34 & 0.8494 & 13.77 & 0.3199  & 27.88 &0.8394 & 30.07 &0.8664\\

  GMM\cite{li2016rain} &29.05 &0.8717 & 15.23 &0.4498  &27.78 & 0.8585 & 32.14 & 0.9145 \\

  JCAS\cite{gu2017joint}  & 28.54 & 0.8524 & 14.62 & 0.4510 &26.20 & 0.8471 & 33.10 &0.9305 \\

  Clear\cite{fu2017clearing} &30.24 & 0.9344 & 15.33 & 0.7421 & 26.21& 0.8951 & 31.24 & 0.9353\\

  DDN\cite{fu2017removing}& 32.38 & 0.9258 & 22.85 & 0.7250 & 28.45 & 0.8888 & 34.04 & 0.9330 \\

  RESCAN\cite{li2018recurrent}   & 38.52& 0.9812 &29.62 & 0.8720 &32.03& 0.9314 &36.43&0.9519\\

  PReNet\cite{ren2019progressive}& 37.45& 0.9790 &30.11& 0.9053 & 32.55 & 0.9459 & 36.66& 0.9610\\

  SPANet\cite{wang2019spatial} & 35.33 & 0.9694 &25.11 & 0.8332 & 29.85& 0.9148 & 35.85& 0.9572 \\

  JORDER\_E\cite{yang2019joint} & 38.59 & 0.9834 & 30.50 &0.8967 &32.00 & 0.9347 & 36.69 & 0.9621\\

  SIRR\cite{wei2019semi} & 32.37 & 0.9258 & 22.47 & 0.7164 & 28.44 & 0.8893 & 34.02& 0.9347\\
  \tr{IDT\cite{xiao2022image}} & \tr{35.42} & \tr{0.9674} & \tr{30.45} & \tr{0.9081} & \tr{\bf 33.55} & \tr{\bf 0.9531} & \tr{35.98} & \tr{0.9584} \\
  \midrule
  RCDNet\cite{wang2020model} & 40.00 & 0.9860 & 31.51 & 0.9119 & 33.10 & 0.9475 & 37.61 & \bf 0.9644 \\
  \tr{RCDNet-$partial$} & \tr{38.02} & \tr{0.9796} & \tr{29.75} & \tr{0.8550} & \tr{31.66} & \tr{0.9305} & \tr{36.58} & \tr{0.9585} \\
  \tr{RCDNet-$pdoe$} & \tr{39.08} & \tr{0.9830} & \tr{30.34} & \tr{0.8979} & \tr{32.92} & \tr{0.9446} & \tr{36.42} & \tr{0.9463} \\
  \tr{RCDNet-$gcnn$} & \tr{39.94} & \tr{0.9856} & \tr{30.88} & \tr{0.9027} & \tr{32.88} & \tr{0.9448} & \tr{37.53} & \tr{0.9618} \\
  \tr{RCDNet-$e2cnn$} & \tr{40.15} & \tr{0.9863} & \tr{31.33} & \tr{0.9090} & \tr{32.18} & \tr{0.9369} & \tr{37.50} & \tr{0.9597} \\
  \tr{RCDNet-$fconv$} & \tr{40.07} & \tr{0.9862} & \tr{30.98} & \tr{0.9038} & \tr{32.66} & \tr{0.9420} & \tr{37.67} & \tr{0.9618} \\
  \tr{RCDNet-$reg$} & \tr{\bf 40.33} & \tr{\bf 0.9868} & \tr{\bf 31.64} & \tr{\bf 0.9138} & \tr{33.49} & \tr{0.9506} & \tr{\bf 37.74} & \tr{0.9633} \\

    \bottomrule
\end{tabular}\vspace{-2mm}
  \vspace{4mm}
  \label{tab:tablederain}
  \vspace{-4mm}
\end{table*}


\begin{figure*}[ht]
    \centering
    \includegraphics[width=17cm]{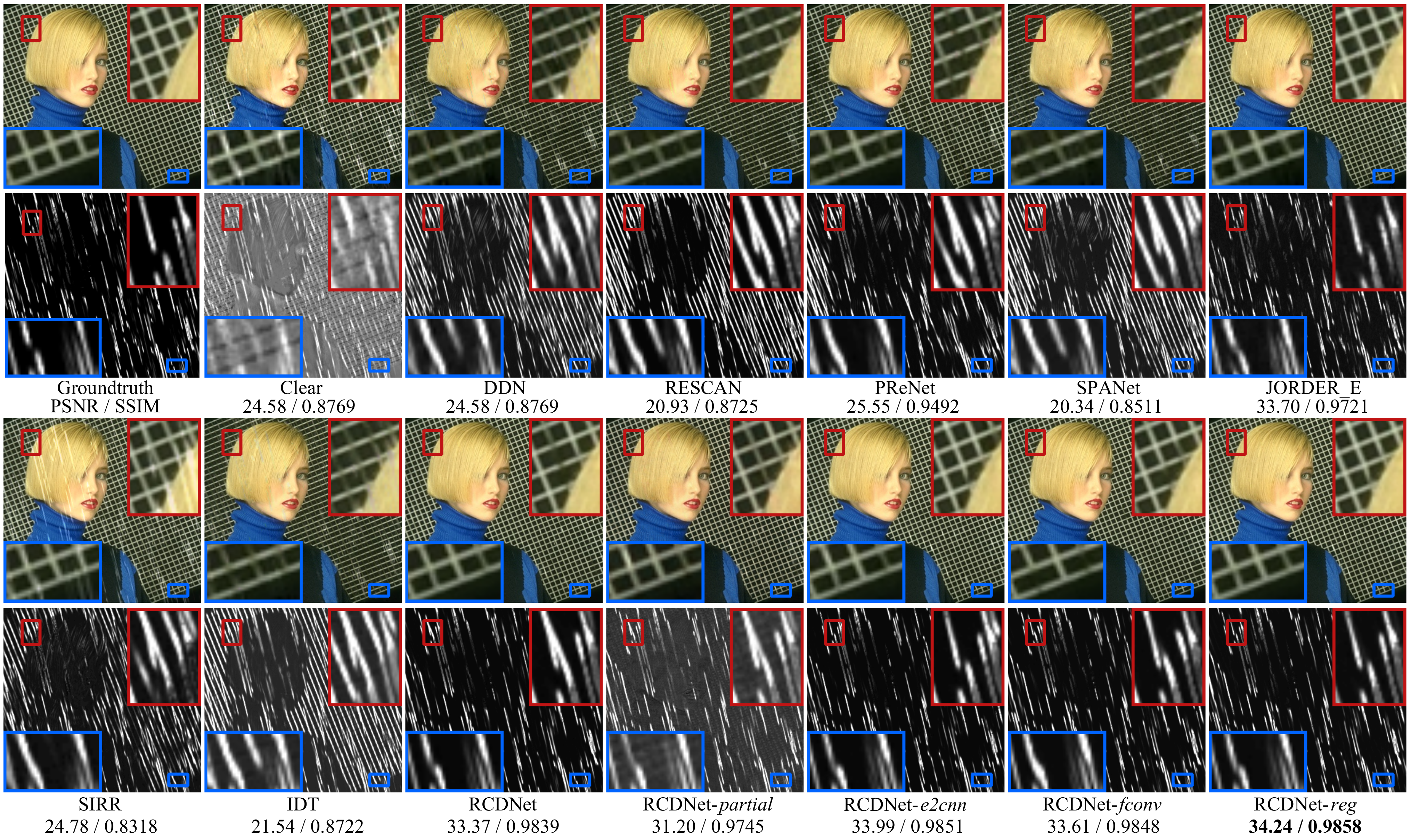}
    \vspace{-2.5mm}
    \caption{The $1^{st}$ column: a typical ground truth sample in Rain100L \cite{yang2019joint} dataset (upper) and its ground truth rain layer (lower). The $2^{nd}-14^{th}$ columns: derained results (upper) and extracted rain layers (lower) by all competing methods}
    \label{fig:derain}
    \vspace{-5mm}
\end{figure*}
\section{Experimental Results} \label{sec:exp}

\subsection{Metal Artifact Reduction in CT Image}

\noindent{\bf Network setting.}
In order to comprehensively verify the effects of the proposed regularizer, We compare the proposed method with existing rotation equivariant methods, including GCNN \cite{cohen2016group}, E2-CNN \cite{cohen2019general}, PDO-eConv \cite{shen2020pdo} and F-Conv \cite{xie2022fourier}, and partial-GCNN \cite{romero2022learning}, the partial equivariant methods. Specifically, for ACDNet, We construct strictly equivariant and partial equivariant network by replacing the regular convolution in the original network with above methods as ACDNet-$gcnn$, ACDNet-$e2cnn$, ACDNet-$pdoe$, ACDNet-$fconv$, and ACDNet-$partial$ respectively. Our method is represented by ACDNet-$reg$. Moreover, we use the same notation strategy for DICDNet and OSCNet, ensuring clarity and coherence throughout. Among these methods, We perform the experiment on the $p4$, rotation group \ie, $t=4$. Besides, the channel numbers of the strictly equivariant methods are set as $\frac{1}{4}$ of the original network. All the training settings and loss function are set the same as the original methods for fair competition.

\vspace{0.8mm}
\noindent{\bf Datasets and Training Settings.} Following the synthetic procedure in \cite{wang2021dicd}, \cite{wang2022OSC}, we can generate the paired $ X$ and $ Y$ for training and testing by using 1,200 clean CT images from DeepLesion \cite{yan2018deeplesion} and 100 simulated metallic implants from \cite{zhang2018convolutional}. Consistent with the original work, we randomly select 90 masks and 1000 clean CT images to synthesize metal-corrupted training samples. The remaining 10 masks and 200 clean CT images as test samples. For training, except the total epoch of DICDNet is changed to 200 to ensure convergence, all methods of replacing the rotation equivariant network remain consistent with the original method.

\vspace{-0.8mm}
\noindent{\bf Quantitative and Qualitative Comparison.}
As illustrated in Table \ref{tab:tabmar}, it is evident that when replacing the networks with its strictly rotation equivariant version, the performance will be slightly reduced, indicating that the symmetry in the data may be imperfect, and the network depicts strict symmetry. Secondly, the performance of rotational equivariant networks implemented by early parameterization methods, such as PDO-eConv and E2CNN, will be significantly degraded due to low representation accuracy. Also, partial-GCNN performs poorly on metal artifact reduction tasks as a way to relax strict equivariant constraints. However, our method can describe the degree of equivariance in the data without damaging the representation accuracy and further outperforms the competing methods. Visually, as demonstrated in Figure \ref{fig:mar}, our method performs better than the original method in removing metal artifacts and restoring the contour of human tissue structures. Both quantitative and qualitative comparisons reveal the superiority of the proposed methods. Additional generalization results can be found in the supplementary material.

\begin{table*}[t]
    \footnotesize
    \renewcommand\arraystretch{1.1}
    \setlength{\tabcolsep}{8.5pt}
    \centering
    \caption{PSNR of pixelwise image inpainting reconstruction for different methods using noisy test measurements.}\vspace{-2mm}
    \begin{tabular}{c c c c c c c c c c c c c}
    \toprule
    $\gamma$ & $A^{\dagger}(y)$ & EI & REI & Partial & PDOe & e2cnn & Fconv & GCNN & EQ-Reg  \\
    \midrule
    0.01 & 5.7 $\pm$ 1.5 & 18.9 $\pm$ 1.0 & 19.7 $\pm$ 2.1 & 13.8 $\pm$ 1.2 & 19.5 $\pm$ 1.5 & 19.7 $\pm$ 1.8 & 20.2 $\pm$ 1.5 & \underline{20.7 $\pm$ 2.0} & \textbf{21.2} $\pm$ \textbf{1.7}  \\
    0.05 & 5.1 $ \pm $ 1.4 & 11.8 $ \pm $ 3.0 & 18.0 $ \pm $ 1.5 & 13.2 $ \pm $ 1.0 & 16.5 $ \pm $ 1.4 & 17.3 $ \pm $ 1.4 & 17.4 $ \pm $ 1.4 & \underline{18.0 $ \pm $ 1.2} & \textbf{18.5} $\pm$ \textbf{1.4} \\
    0.1 & 4.4 $\pm$ 1.3 & 9.8 $\pm$ 0.8 & \underline{16.6 $\pm$ 0.2} & 12.3 $\pm$ 1.1 & 15.0 $\pm$ 1.3 & 15.4 $\pm$ 1.3 & 15.7 $\pm$ 1.1 & 16.5 $\pm$ 1.5 & \textbf{16.9} $\pm$ \textbf{1.3}  \\
    \bottomrule
    \end{tabular}
    \label{tab:inpainting}\vspace{-2mm}
\end{table*}

\begin{figure*}[ht]
    \centering
    \includegraphics[width=17cm]{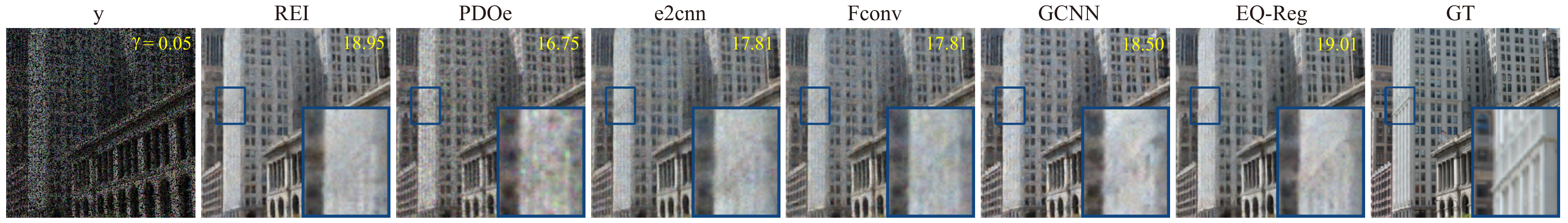}
    \vspace{-3mm}
    \caption{Inpainting reconstructions on test images with Poisson noise ($ \gamma$ = 0.05) and 30\% mask rate. PSNR values are shown in the top right corner of the images.}
    \label{fig:inpainting}
    \vspace{-5mm}
\end{figure*}


\vspace{-1mm}
\subsection{Single Image Rain Removal} \label{sec:exp_derain}
\vspace{-0.5mm}
\noindent{\bf Network setting.} Similar to the experiments in Section 4.1, we construct RCDNet-$gcnn$, RCDNet-$e2cnn$, RCDNet-$pdoe$, RCDNet-$fconv$, RCDNet-$partial$ to verify the effectiveness of strictly rotation equivariant networks, with setting $t = 2$ \& $\frac{1}{2}$ channel number. Because in the current rain removal task, the rain strip types in the datasets usually present a 180-degree rotational symmetry relationship. We therefore expect the network to have two angles, that is, an approximate equivariant of 180 degrees. All the training settings and loss function are set the same as the original RCDNet for fair competition.

\noindent{\bf Datasets and Training settings.} We compare our method with typical single image rain removal SOTA methods, including DSC~\cite{luo2015removing}, GMM~\cite{li2016rain}, JCAS~\cite{gu2017joint}, Clear~\cite{fu2017clearing}, DDN~\cite{fu2017removing}, RESCAN~\cite{li2018recurrent}, PReNet~\cite{ren2019progressive}, SPANet~\cite{wang2019spatial}, JORDER\_E~\cite{yang2019joint}, SIRR~\cite{wei2019semi}, and RCDNet on four commonly-used benchmark datasets, \ie, Rain100L \cite{yang2019joint}, Rain100H \cite{yang2019joint}, Rain1400 \cite{fu2017removing}, and Rain12~\cite{li2016rain}. The training strategy is carried out according to the original setting.

\noindent{\bf Quantitative and Qualitative Comparison.}
As shown in Table \ref{tab:tablederain}, with the proposed equivariant regularization, RCDNet-$reg$ consistently outperform the method of strict or partial equivariant RCDNet, and achieve comparable performance with the transformer-based SOTA method IDT. Particularly, in the datasets Rain100L, Rain100H and Rain12, the RCDNet-$reg$ method achieves the best performance among these methods. In addition, as shown in Figure \ref{fig:derain}, our proposed method exhibits superior performance in removing visual rain streaks without compromising the details of the original image. It is worth noting that the model based on convolution sparse coding often mistakenly identifies white stripes as rain stripes, while the addition of equivariant regularization can effectively alleviate this shortcoming. These results support the effectiveness of adopting equivariant regularization in this task.

\vspace{-1mm}
\subsection{Image Inpainting}
\vspace{-1mm}

\noindent {\bf Experiment Setting.} We adopt the self-supervised framework for learning from noisy and partial measurements introduced in EI \cite{chen2021equivariant} and REI \cite{chen2022robust}, reproducing the original image inpainting experiment as described in these works. Building upon this setup, we compare the methods of strict equivariant and partial equivariant with the methods presented in this paper. Specifically, we assess reconstruction performance on the Urban100 dataset \cite{huang2015urban} by applying a random mask covering $30\%$ of the image and performing restoration under various Poisson noise levels. All training configurations and the loss function are maintained as in \cite{chen2022robust}.

\noindent {\bf Quantitative and Qualitative Comparison.} As shown in Table \ref{tab:inpainting}, the proposed method achieves the best effect under different noise intensity interference. The visualization results in Figure \ref{fig:inpainting} also show that the proposed method can better remove noise and restore original image details. Additional visualizations and further analysis can be found in the supplementary material.

\vspace{-1mm}
\subsection{Image classification}
\vspace{-1mm}
We compare the proposed method with the existing partial equivariant approach, Partial-GCNN \cite{romero2022learning}, on the CIFAR-100 image classification task. Table \ref{tab:classification} indicates that our approach consistently outperforms the competing methods.
\vspace{-2mm}

\begin{table}[h]
\centering
\footnotesize
\captionsetup{justification=centering}
\setlength{\tabcolsep}{3.5pt}
\caption{Classification accuracy of CIFAR100 dataset.}\vspace{-3mm}
\begin{tabular}{>{\centering\arraybackslash}p{1.2cm}  
                    >{\centering\arraybackslash}p{1.1cm}
                    >{\centering\arraybackslash}p{1cm}
                    >{\centering\arraybackslash}p{0.8cm}
                    >{\centering\arraybackslash}p{1.7cm}
                    >{\centering\arraybackslash}p{1cm}}
\toprule
Dataset & Group & No.elems & GCNN & Partial-GCNN & EQ-Reg \\
\midrule
\multirow{2}{*}{CIFAR100} & \multirow{2}{*}{SE(2)} & $t=4$ & 49.79 & 53.28 & \textbf{56.10} \\
 & & $t=8$ & 53.98 & 56.53 & \textbf{57.56} \\
\bottomrule
\end{tabular}
\label{tab:classification}
\vspace{-6mm}
\end{table}

\section{Conclusion}\vspace{-2mm}
In this paper, we introduce a simple yet effective regularizer-based equivariant strategy for image processing tasks, capable of adaptively adjusting equivariance to suit various applications. 
Extensive experiments across multiple tasks validate the superior performance and generalizability of the proposed method, surpassing current SOTA methods. Moreover, this approach paves the way for developing equivariant networks beyond the conventional CNN framework, expanding the possibilities for future network design, including the equivariant Transformer-based networks and networks equivariant with respect to other transformations.

\footnotesize
\noindent\textbf{Acknowledgment}. This research was supported by the NSFC project under contract U21A6005, the Major Key Project of PCL under Grant PCL2024A06, the Tianyuan Fund for Mathematics of the National Natural Science Foundation of China (Grant No. 12426105), and the Key Research and Development Program (Grant No. 2024YFA1012000).

{
    \small
    \bibliography{main}
}

\end{document}